\pdfoutput=1

\documentclass[11pt]{article}

\usepackage{naacl2021}

\usepackage{times}
\usepackage{latexsym}

\usepackage[T1]{fontenc}

\usepackage[utf8]{inputenc}

\usepackage{microtype}

\title{Introducing Information Retrieval for Biomedical Informatics Students}

\author{
  Sanya B. Taneja \ \ \ \ \ \ 
  Richard D. Boyce \ \ \ \ \ \ 
  William T. Reynolds \ \ \ \ \ \
  Denis Newman-Griffis 
  \\
  University of Pittsburgh \\
  Pittsburgh, PA, USA\\
  \texttt{\{sbt12, rdb20, wtr8, dnewmangriffis\}@pitt.edu}
}

\begin{document}
\maketitle

\begin{abstract}
Introducing biomedical informatics (BMI) students to natural language processing (NLP) requires balancing technical depth with practical know-how to address application-focused needs. We developed a set of three activities introducing introductory BMI students to information retrieval with NLP, covering document representation strategies and language models from TF-IDF to BERT. These activities provide students with hands-on experience targeted towards common use cases, and introduce fundamental components of NLP workflows for a wide variety of applications.

\end{abstract}

\section{Introduction}
\label{sec:introduction}

Natural language processing (NLP) technologies have become a fundamental tool for biomedical informatics (BMI) research, with uses ranging from mining new protein-protein interactions from scientific literature to recommending medications in clinical care. In most cases, however, the research question at hand is a biological or clinical one, and NLP tools are used off-the-shelf or lightly adapted rather than being the focus of research and development. When introducing BMI students to NLP, instructors must therefore navigate a balance between the computational and linguistic insights behind why NLP technologies work the way they do, and practical know-how for the application-focused settings where most students will go on to use NLP.

We developed a set of three activities designed to expose introductory BMI students to the fundamentals of NLP and provide hands-on experience with NLP techniques. These activities were designed for use in the Foundations of Biomedical Informatics I course at the University of Pittsburgh, a survey course which introduces students to core methods and topics in biomedical informatics. The course is required for all students in the Biomedical Informatics Training Program; students have a range of experience in computer science, and no background in artificial intelligence or NLP is required. The sequence of activities, implemented as Jupyter notebooks, comprise a single assignment focused on information retrieval, a common use case for NLP in all areas of BMI research. The assignment followed lectures focused on information retrieval, word embeddings, and language models (presented online in Fall 2020). \S\ref{sec:assignment-details} gives an overview of the three activities, and we note directions for further refinement of the activities in \S\ref{sec:discussion}.

\section{Assignment Details}
\label{sec:assignment-details}

Our set of three activities was designed with two primary learning goals in mind:
\vspace{-0.2cm}
\begin{itemize}
    \itemsep0em
    \item Expose introductory BMI students to fundamental strategies for text representation and language models, geared towards information retrieval in biomedical contexts; and
    \item Provide students with hands-on experience creating NLP workflows using pre-built tools.

\end{itemize}\vspace{-0.2cm}
Our Jupyter notebooks provide a sequence of code samples to analyze and execute, combined with background questions to assess understanding of the computational and linguistic insights behind the NLP technologies students are using.

{\bf Notebook 1: Fundamentals of document analysis.}
In this notebook, students were first introduced to basic preprocessing tasks in NLP workflows such as tokenization, stemming, casing, and stop-word removal. Using a corpus from the Natural Language Toolkit (NLTK) \cite{bird2009natural}, the notebook demonstrated two indexing techniques - inverted indexing and creation of a weighted document-term matrix using term frequency-inverse document frequency (TF-IDF). Students then implemented a synthetic information retrieval task involving a collection of 12 documents mapped to 20 queries as a reference set. The students evaluated the information retrieval system with synthetic results for two queries comprising numeric values for each document in the query. Students measured system performance using TREC evaluation measures including recall, precision, interpolated precision-recall average, and mean average precision. The evaluation measures were implemented using the pytrec\_eval library \cite{VanGysel2018pytreceval}.

{\bf Notebook 2: Introduction to word embeddings.}
In this notebook, students were introduced to word embeddings as a text representation tool for NLP. The students first created embeddings using singular value decomposition (SVD) of a co-occurrence matrix using the corpus from Notebook 1. This established the idea of capturing semantic similarity in texts as opposed to a tradition bag-of-words model. The notebook then used pretrained word2vec embeddings \cite{mikolov2013distributed} to demonstrate a more refined approach of SVD. The students were able to visualize both the embedding approaches in the notebook. This was particularly important as most students did not have prior experience with embeddings and plotting the word embeddings can lead to greater insight into the variable semantic similarity that embedding representations provide over lexical features.

{\bf Notebook 3: Introduction to BERT and clinicalBERT.}
As all the activities are designed for introductory students, YouTube tutorials were used to provide background on neural networks and design decisions in language models for students with minimal background in NLP. Students were introduced to NLP workflows and language models using the Transformers library. Transformers is an open-source library developed by Hugging Face that provides a collection of pretrained models and general-purpose architectures for natural language processing \cite{wolf-etal-2020-transformers}, based on the Transformer architecture \cite{Vaswani2017}. This includes BERT \cite{Devlin2019} and clinicalBERT \cite{alsentzer-etal-2019-publicly}, which are used in this notebook to implement Named Entity Recognition and Medical Language Inference tasks.

The notebook guided the students through a Medical Language Inference task to infer knowledge from clinical texts. Language inference in medicine is an important application of NLP as clinical notes such as those containing past medical history of a patient contain vital information that is utilized by clinicians to draw useful inferences \cite{romanov-shivade-2018-lessons}. The task also introduced BMI students to challenges unique to applying NLP on clinical texts such as domain-specific vocabulary, diversity in abbreviations, content structure, and named entity recognition with clinical jargon. The students used the MedNLI \cite{romanov-shivade-2018-lessons} dataset created from MIMIC III \cite{johnson2016mimic} clinical notes for this task. Building on knowledge from the previous notebooks, they implemented workflows to compare the performance of BERT and clinicalBERT models for prediction of labels in clinical texts. The students were encouraged to understand the importance of domain representation in pretraining data and the process of fine-tuning NLP models for domain-specific language.
\section{Discussion}
\label{sec:discussion}

We designed three activities to demonstrate fundamental concepts and workflows for application of NLP to introductory BMI students. While the scope of NLP in the biomedical field is much larger than one assignment, we developed the activities to provide students with a modular workflow of components that are applicable to other NLP applications besides information retrieval; i.e., text preprocessing, indexing, execution, and evaluation.

One practical challenge for clinically-focused exercises is the limited availability of benchmark datasets. Most clinical datasets require Data Use Agreements and individual training requirements that are cumbersome for a classroom setting.Thus, the first two notebooks use the NLTK corpus which is a popular dataset for introducing NLP concepts without the challenges present in clinical datasets. While MIMIC \cite{johnson2016mimic} is a valuable, relatively accessible source for clinical text, available annotations for it are limited. Students can thus be introduced to fine-tuning of language models for the specifics of medical language, but instructors must anticipate challenges in providing train/test splits for supervised machine learning.

We further take advantage of popular pre-built libraries (like Transformers) so that students can focus on the application rather than constructing neural networks. In an application-focused setting, technical knowledge of neural NLP systems is less necessary, but users of those systems still need to understand what kinds of regularities they rely on and when they may be unreliable. The activities are thus designed to reflect the perspective of the practical challenges that students will face when working in biomedical NLP.

Finally, one important area we are investigating as we refine and extend these teaching materials is the ethical considerations of biomedical AI technologies. The intersection of medical and AI ethics poses several challenging questions for designing, training, and applying AI technologies in the biomedical setting \cite{Char2018,KESKINBORA2019277}, and sensitive information often described in medical text presents further ethical questions for NLP systems \cite{lehman2021does}. Thus, determining what BMI students have a responsibility to understand about the NLP tools they use, and how we most effectively teach that information in limited course time, is key to broadening the responsible use of NLP in BMI research.

All materials presented in this paper are available from \url{https://github.com/dbmi-pitt/bioinf\_teachingNLP}.

\section*{Acknowledgments}
This work was supported in part by the National Library of Medicine of the National Institutes of Health under award number T15 LM007059.

\bibliography{references}
\bibliographystyle{acl_natbib}

\end{document}